\documentclass{birkjour_t2}
 \newtheorem{thm}{Theorem}[section]

 \theoremstyle{definition}
 
 \theoremstyle{remark}
 
 \newtheorem{example}[thm]{Example}
  \newtheorem{PS}[thm]{Proof Strategy}
    
 \numberwithin{equation}{section}

\usepackage{hyperref}
\usepackage{tikz}
\usetikzlibrary{decorations.markings}
\usetikzlibrary{arrows}
\usepackage{amssymb}
\usepackage{amsfonts,amsmath,latexsym,stmaryrd}
\usepackage{listings}

\newcommand{\C}[1]{\mbox{\lstinline`#1`}}

\let\L=\lstinline

\definecolor{dkblue}{rgb}{0,0.1,0.5} 
\definecolor{lightblue}{rgb}{0,0.5,0.5} 
\definecolor{dkgreen}{rgb}{0,0.4,0} 
\definecolor{dk2green}{rgb}{0.4,0,0} 
\definecolor{dkviolet}{rgb}{0.6,0,0.8}
\definecolor{shadethmcolor}{rgb}{0.9, 0.9,1}

\usepackage{verbatim}

\usepackage{arydshln}

\usepackage{cite}

\usepackage{xcolor,colortbl}

\begin{document}
%-------------------------------------------------------------------------
% editorial commands: to be inserted by the editorial office
%
%\firstpage{1}
%\volume{228}
%\DOI{003-0001}
%\seriesextraline{This is the Concrete Title of this Book\br H.E. R and S.T.C. W, Eds.}
%\firstpage{1}
\title[Recycling Proof Patterns in Coq: Case Studies]{Recycling Proof Patterns in Coq: Case Studies}
%----------Author 1
\author[J. Heras]{J\'onathan Heras}

\address{%
School of Computing, University of Dundee, UK}

\email{jonathanheras@computing.dundee.ac.uk}

\thanks{The work was supported by EPSRC grant EP/J014222/1 and EP/K031864/1}
%----------Author 2
\author[E. Komendantskaya]{Ekaterina Komendantskaya}
\address{School of Computing, University of Dundee, UK}
\email{katya@computing.dundee.ac.uk}


\date{}

\begin{abstract} 
Development of Interactive Theorem Provers has led to the creation of big libraries and varied infrastructures for formal proofs. 
However, despite (or perhaps due to) their sophistication, the re-use of libraries by non-experts or across domains is a challenge. 
In this paper, we provide detailed case studies and evaluate the machine-learning tool ML4PG built to interactively data-mine the
electronic libraries of proofs, and to provide user guidance on the basis of proof patterns found in  the existing libraries. % or libraries being written by other users.   
%In this

\end{abstract}

%%% ----------------------------------------------------------------------
\maketitle
%%% ----------------------------------------------------------------------

\section{Introduction}

%Donald Knuth famously compared computer programming to an art \cite{Knuth74}. % in which some insight is required to combine a fixed number 
%of  programming language %(no ``the'' because of 59B)
% constructs into a valid and elegant program. %Thus, programming skills depend on experience and correct intuition.
%This comparison still holds for 
Interactive theorem provers (ITPs)  (e.g. ACL2~\cite{KMM00-1}, Agda~\cite{Agda}, Coq~\cite{Coq}, Isabelle/HOL~\cite{NPW02}, Matita~\cite{Matita}, Mizar~\cite{mizar} to name a few) are a family of 
systems
%offering a rich enough syntax to 
allowing the formalisation  of a wide variety of domains, ranging from  mathematical theories to software verification. % of practically unlimited sophistication and complexity. 
The most recent achievements concerned formalisation and computer verification of results coming from  Group Theory~\cite{FTT}, Real Numbers~\cite{realsCoq}, 
Discrete Mathematics~\cite{flyspeck} and Security~\cite{ACDDHPPPT14}.
The successful and efficient ITP programming 
%This 
often requires a combination of mathematical and programming intuition; see e.g. \cite{Ben06}.
There can be a rich variety of approaches to the formalisation and proof development for a given task. %; and the language  generally offers little guidance in the choice of such directions.  
 Thus, a programmer 
relies on his previous experience and ability to \emph{``creatively''} adapt
already used proof techniques and \emph{patterns} in newly constructed proofs. The situation is similar in mathematical proofs that do not use ITPs; however, 
the low-level details of ITP proofs and the peculiarities of these systems make the discovery of patterns and proof techniques more difficult. 
This explains why a ``steep learning curve'' is often mentioned as one of the big obstacles to wider adoption of ITPs. 
% --
% it takes several months for users to become proficient using ITPs and also to learn the techniques and patterns that can be applied in different proofs. % by professional mathematicians or 
%industries alike.
In this paper, we are probing the abilities of our recent machine-learning tool ML4PG~\cite{HK12,HK14,KHG13} to find interesting proof patterns automatically, and thus enable a more efficient use of ITPs by specialists coming from a wider range of domains.

The development of ITPs has led to the creation of big libraries and varied infrastructures for formal
mathematical proofs. 
These frameworks usually involve thousands of definitions and theorems (for instance, there are approximately
4200 definitions and 15000 theorems in the formalisation of the Feit-Thompson theorem~\cite{FTT}). % Definitions,
%theorems and proof-strategies used in
Parts of  those libraries can often be re-applied in new domains; however, 
it is a challenge for expert and non-expert users alike to trace them and find re-usable concepts and proof ideas.

A different, but related, challenge is faced during the creation of these libraries. These frameworks are developed by teams
(e.g. 15 people were involved in the Feit-Thompson theorem project), and the situation is similar in industry where teams use ITPs to verify 
the correctness of hardware and software systems. % -- in the industrial context, proofs usually have more regularity and involve
%many routine or similar cases. 
In those teams, each user has his own definitions, notation and proof-style, which makes the 
collaborative proof development difficult. In both scenarios, it would be extremely helpful to use a tool that could detect patterns across different users, notation and libraries.

%In this paper, we are making the first steps towards automated detection of significant proof patterns in Coq/SSReflect. 
% As a solution, 
To address these challenges, we propose ML4PG -- a machine-learning extension to the Proof General~\cite{ProofGeneral} interface for Coq~\cite{BC04} and 
its SSReflect dialect~\cite{SSReflect}. 
%Coq/SSReflect is the higher-order interative theorem prover from a wider range of similar tools, e.g. Isabelle, HOL, AGDA, ACL2.
Our main goal is to prove the concept:
\emph{it is possible to embed a lightweight statistical machine-learning tool into an ITP proof interface, and use it interactively to find non-trivial patterns in existing proofs
and thus aid new proof developments}.
 %To our knowledge, our work is the first attempt to incorporate some machine-learning practices into Coq/SSReflect proof development.

%We have described, in a companion paper \cite{KHG13}, ML4PG -- an extension to Proof General 
% that automatically clusters Coq/SSReflect proofs. % based on statistics of low-level proof features. %patterns arising during the interactive proof development.
%In brief, ML4PG does the following:
The ML4PG package for Proof General features the following main functions:

\begin{itemize}
\item The user works within the interactive environment of Coq/SSReflect, 
and has an option to call ML4PG from the Proof General interface whenever he needs to see similar proof patterns. 
%receive 
%some statistical machine-learning hint.

\item Based on the user's choice, ML4PG compiles the chosen libraries, and extracts significant proof features from the existing lemmas and proofs;

\item ML4PG connects to machine-learning tools, and runs a number of experiments on clustering the data for each user query. Based on the results, it chooses the most reliable patterns;
thus relieving the Coq programmer of the laborious step of post-processing the statistical results.

\item If the user chooses to see only patterns related to his current proof goal, ML4PG would further filter the results  and show the families of related proofs to the user.  

\item There are two ways in which we can read the results. The related theorems/lemmas are by default displayed in a separate window. Additionally, ML4PG can display the proof pattern in  the form of an automaton,
showing the correlating features that made the pattern. 
%Pre-processing existing proofs: the user may choose to find proof-patterns in existing libraries; the patterns may be based on similarities of definitions and lemma statements; or on the similarities   in which case, ML4PG clusters 
%	\item[\textbf{F1.}] it works on the background of Proof General, and extracts some simple, low-level features from interactive proofs in Coq/SSReflect;
%	\item[\textbf{F2.}] it automatically sends the gathered statistics to a chosen machine-learning interface and triggers execution of a clustering algorithm of the user's choice;
%	\item[\textbf{F3.}] it does some gentle post-processing of the results given by the machine-learning tool, and displays families of related proofs to the user.
	\end{itemize}

%\begin{itemize}
%	\item For every proof in a selected set of libraries, it extracts a number of low-level proof features;
%	\item It then sends this data to a  state-of-the-art statistical machine-learning interface chosen by the user. Currently, MATLAB and Weka are the two options available in ML4PG;
%	\item ML4PG allows the user to choose a particular statistical algorithm available in the machine-learning interface, as well as to fix certain 
%	learning parameters, see also Section~\ref{sec:ML4PG};
%	\item The machine learning engine then classifies the available proofs into small or big clusters (depending on the learning parameters above), 
%	and displays the clusters with lemma names back in Proof General.
%\end{itemize}
 
Section~\ref{sec:ML4PG} gives an overview of ML4PG features, details of its implementation are given in~\cite{KHG13}. ML4PG's features have been substantially extended since \cite{KHG13},  a detailed description of the new features is given in~\cite{HK14}. 
In this paper, we do not focus on the ML4PG implementation \emph{per se}, although we use it for all the examples and experiments shown in this paper.
Our main goal here is to show how useful the automated proof pattern detection can be in different domains.
%In this regard, we pose the following three questions:
 %\begin{enumerate}
% \item Are there automatically detectable proof patterns in proofs developed in Coq/SSRefelect?
% \item Where do these patterns arise and how can they be statistically detected?
% \item How can we use them to make theorem proving more efficient?
%\end{enumerate}  
%We answer these  questions by 

To illustrate this, we devise three experiments (``user scenarios'') to test ML4PG. 
Each example is designed to demonstrate a different 
aspect of proof-pattern recognition.  To demonstrate ML4PG's ability to adapt to different domains, 
we deliberately illustrate each user scenario by
 using libraries coming from different subject areas, ranging from basic mathematical infrastructures to software verification. 

\textbf{User Scenario 1} illustrates how to use ML4PG for detecting proof patterns prior to the start of a new proof development. To achieve this,
Section~\ref{sec:preproof} analyses fundamental libraries that are common in most developments using the SSReflect library~\cite{SSReflect}.
The SSReflect library was developed as the infrastructure for the formalisation of the Four Colour Theorem~\cite{FCT} and has played 
a key role in the formal proof of the Feit-Thompson theorem~\cite{FTT}. Up to version 1.4, the SSReflect library was distributed together 
with the MathComp library (that contains the theories about the development of the proof of the Feit-Thompson theorem); from version 1.5, the SSReflect
library can be downloaded independently from the MathComp library.

In this first scenario, we use pattern recognition with  the aim of spotting common proof patterns across fundamental libraries (1404 theorems). 
The benefits of using ML4PG in this context is that it can be used to speed up the beginning of a proof development, making it easier to recycle
patterns already available in the libraries.

\textbf{User Scenario 2} considers the problem of proof-pattern discovery in a different light. In User Scenario 1, there was always an interesting underlying 
proof pattern hidden in the big proof libraries, ``waiting'' to be discovered. What if, despite the user's hope that one library may contain similar proof strategies 
to another, the actual proofs are in fact too different to be recycled? 
Section~\ref{sec:nash} studies the results that ML4PG obtains working with two different Coq libraries formalising results from  game theory~\cite{Ves06,nash}.
One might hope that they contain similar proof patterns, since they formalise  the same subject domain; but in fact, ML4PG shows that the actual proof strategies used in
\cite{nash} and \cite{Ves06} are completely different. 
%It is the opposite situation to User Scenario 2, where common proof patterns were found despite the fact that the libraries were coming from three different domains. 
This ``negative'' output given by ML4PG may in reality save the user's time inspecting these libraries manually.
% In the scenario 4, 
%we analyse  proof patterns arising between lemmas of the same library and also across libraries.

\textbf{User Scenario 3} considers the situation when a team of several people develops  a set of different modules within one bigger (industrial-scale)
verification effort, see Section \ref{sec:JVM}. 
%Thus, 
%Section \ref{sec:JVM} takes one further step from academic to industrial applications of Coq and ML4PG. 
For this purpose, we 
translate  the proofs of correctness of the Java Virtual Machine (JVM) given in~\cite{M03} into Coq. The industrial scenario of interactive theorem proving may 
 differ significantly from the scenarios above. Namely, industrial verification tasks often feature a bigger number of routine cases and similar lemmas;
and also such tasks are distributed across a team of developers. Here, the inefficiency of automated proving often arises when programmers use different notation to 
accomplish very similar tasks, and thus a lot of work gets duplicated, see also~\cite{BHJM12}.
We tested ML4PG in exactly such a scenario: we assumed that a programming team has collectively developed proofs of 
%\begin{itemize}
\emph{a) soundness of specification, and b) correctness of implementation}
%\end{itemize}
of Java byte code
 for a dozen of programs computing multiplication, powers, exponentiation, and other arithmetic functions. %; each containing around 12 lemmas and definitions.
We assumed that there is a relative novice in the team, trying to ``learn'' from the previous team efforts, in order to repeat their proof steps for a new Java function (factorial in our setting).
He calls ML4PG, which discovers common patterns among these proofs and relevant lemmas (around 150 training examples in total). The suggested clusters indeed helped
to advance the proofs of properties a) and b) for the Java byte code of the factorial function.   
%that is to say, we have defined an interpreter for JVM programs within this proof 
%assistant. From now on, we refer to our machine as ``Coq JVM'' or ``CJVM''. 
%This work is inspired by a series of JVM models formalised in ACL2~\cite{M99,M03,LM03}.

This is the first detailed evaluation of ML4PG, note that \cite{KHG13} focused mainly on the user interface and contained very simple examples.
The case studies presented here convince us that when ML4PG statistically discovers \emph{proof clusters}, it does actually find meaningful, non-trivial and interesting patterns in proofs across different libraries, theories and users. This kind of proof analysis can facilitate the use of Coq to domain expert users, see Section~\ref{expert}.
ML4PG works on the background of Proof General, and if called, provides clustering results almost instantly; thus, can be used interactively, as a handy tool on request.
%In this sense, statistically found clusters and proof patterns essentially provide automatically generated proof hints. 
%providing non-trivial suggestions about analogies between fragments of apparently unrelated libraries, and 
%detecting common patterns in proofs developed by a team.
Finally, it may  be used for educational purposes, as automated proof-pattern recognition may help to smooth the learning curve, see User Scenario 3. 

 ML4PG and all examples presented in this paper are available in \cite{HK12}. 

\textbf{Related Work.}
ML4PG's originality is two-fold, as it can be compared to alternative methods of using machine-learning in automated theorem proving, as well as to Coq/SSReflect tools
allowing interactive pattern-search. 

Related work on using machine-learning in ITPs 
concerned hints in lemma generation for Isabelle/HOL \cite{JDB11}, proof strategy discovery in Isabelle/HOL \cite{BB05}, speed up in proof automation in Mizar \cite{ku12} 
and statistical tactic analysis  in Isabelle~\cite{Duncan02}.
Comparing to these tools, we use unsupervised, rather than supervised, learning; and we do not use sparse  machine-learning methods.
(See also \cite{KHG13} for a detailed comparison of different machine learning tools applied in various theorem provers.)
%Unlike related work on using machine-learning for automated premise selection \cite{ku12,Mash},
 We do not have a
quantitative target when it comes to improving \emph{interactive} proof building experience: 
neither speed up in automated proof search nor the number of automatically proven theorems are the main criteria of success.
Instead, the user experience is the main parameter we target.
We generally follow the ``qualitative'' intuition that ML4PG, being an interactive hint generator, must provide interesting and non-trivial hints on user's demands,
and should be flexible and fast enough to do so in real  time,  at any stage of the proof, and relative to any chosen proof library. 
%ML4PG puts emphasis on improving user interaction with the prover, rather than on proof automation; and 

 Comparing to some of the above approaches, ML4PG does not only analyse the lemma statements, but also involves user tactics and user-defined proof-steps 
into the statistical proof-pattern recognition process. 
This feature also makes ML4PG sensitive (or \emph{adaptable}) to proof styles
innate to a particular user, research community, or subject area (cf. Sections~\ref{sec:preproof}--\ref{sec:JVM}).
In illustration of this point, User Scenario 3 and Section~\ref{sec:JVM} consider cases when different lemma statements have similar proofs; 
User Scenario 2 and Section~\ref{sec:nash} discuss cases when similar lemma statements require a completely different proof strategy.

Comparing to symbolic methods of proof pattern search in Coq/SSReflect, 
%Its main purpose is to assist interaction between the programmer and 
%the ITP, helping the user to find similar proofs across libraries. In this sense, 
%ML4PG extends the functionality of the existing tools for 
%intelligent proof search in Coq: 
e.g. \lstinline?SearchPattern?, \lstinline?Search?, \lstinline?SearchAbout?, \lstinline?SearchRewrite? \cite{Coq,SSReflect}
and \texttt{Whelp} \cite{AspertiGCTZ04}, ML4PG's originality is in introducing statistical pattern-recognition into the rich family of existing searching mechanisms in Coq/SSReflect.
Unlike symbolic pattern-search, ML4PG can discover ``unexpected'' 
proof-patterns that go beyond the patterns 
the user would try as a searching template when using symbolic pattern-search facilities.
Whereas the existing Coq searching tools try to match the user-provided template with other lemma/theorem \emph{statements}, ML4PG takes into consideration the \emph{proof} statistics in conjunction with the lemma shapes.
These two features -- pattern-search without a pre-defined template and the attention to the various proof parameters -- allow to achieve results often orthogonal to
the symbolic pattern search. 

%n some scenarious -- especially when the user is not sure of the pattern he is 

%The rest of the paper is structured as follows.
%Section \ref{sec:interface} gives a brief introduction to ML4PG's facilities from the user's perspective, and thereby reminds of the interactive proof-pattern recognition cycle 
%developed in \cite{KHG13}.
%Section \ref{sec:ml4pg} is devoted to technical description of methods developed in the Stage-2 of ML4PG development, that is, methods of type and definition clustering,
%recurrent symbol clustering, proof-patch method and dynamic lemma numbering.
%Section \ref{sec:scenarios} tests ML4PG on three most common user scenarios: data-mining an existing proof-library prior to the new proof development, goal-directed
%proof-pattern recognition involving several mathematical libraries, and industrial scenario of data-mining proofs across several developers.
%Section \ref{sec:search-coq} gives insight into various methodological decisions made in the design of ML4PG, by comparing ML4PG with existing pattern-search methods 
%in Coq/SSReflect; and comparing ML4PG performance against alternative implementation decisions. Finally, Section \ref{sec:conclusions} concludes the paper.    

\section{Overview of ML4PG}\label{sec:ML4PG}

In this section, we present the main functionality that ML4PG offers to the user. ML4PG works with Coq and its SSReflect dialect, 
and it does not assume any machine-learning knowledge from the user.
The  guidance it provides may come in different forms.
%\begin{enumerate}
%\item the user may prefer to see similar definitions and lemma statements;
%\item the user may prefer to see similar proof strategies (involving the proof tactics);
%\item 
The user may prefer the statistical hint to be related to the current proof-step (cf. User Scenario 3), or give information about proof-patterns arising in a library 
irrespective of the current proof-step (cf. User Scenarios 1 and 2).
The user may choose to data-mine only the current library, or a number of proof-libraries coming from different domains or different users. 
Finally, the user may wish to experiment with proof clusters of different sizes or with different machine-learning algorithms, see Table~\ref{tab:jvm}.
%\end{enumerate}
These choices are accommodated within ML4PG, see \cite{KHG13} for a detailed description of the user interface.
 
%ML4PG accommodates 2 buttons in the Proof General interface that can be used respectively to: (1) find similar proofs in relation to an unfinished proof and 
%(2) find families of similar proofs irrespective of the current proof step. %, (3) find definitions that are similar to a concrete definition, and (4) find families of similar definitions. 
ML4PG functionality is achieved in the following way.

\begin{itemize}
 \item[\textbf{F.1.}] It works on the background of Proof General extracting some low-level features from proofs in Coq/SSReflect.
 \item[\textbf{F.2.}] It automatically sends the gathered statistics to a chosen machine-learning interface and triggers execution of a 
 clustering algorithm according to the choice of the user.
 \item[\textbf{F.3.}] It does some post-processing of the results given by the machine-learning tool, and displays families of 
 related proofs to the user; on request, it shows an automaton explaining which proof features determined the pattern.
\end{itemize}

Stage \textbf{F.1} is devoted to collecting statistics from proofs. 
The discovery of statistically significant features in data is a
research area of its own in machine-learning, known as \emph{feature extraction}, see~\cite{Bishop}. Statistical machine-learning algorithms 
classify given examples seen as points in an $m$-dimensional space, where $m$ is the maximum number of features each example may be characterised by. 
Irrespective of the particular feature extraction algorithm used, most pattern recognition tools~\cite{Bishop} will require that the number of selected
features is limited and fixed -- the exception to this is a special class of methods called ``sparse'' methods~\cite{Blum92}.

ML4PG has its own feature extraction method that collects statistics from the interaction between the user and  the prover. % (recording both goals and tactics). 
The feature extraction is done at the time of the interactive proof construction in the current library or during the Coq compilation for an external library. 
The feature extraction method %implemented in ML4PG 
captures information from proofs 
based on the correlation of a few chosen parameters within five proof steps. For each proof step, the parameters are:

\begin{itemize}
\item[1-2] the names and the number of tactics used in one command line, 
\item[3] types of the tactic arguments;
\item[4] relation of the tactic arguments to the (inductive) hypotheses or library lemmas, 
\item[5-7] three top symbols in the term-tree of the current subgoal, 
and 
\item[8] the number of subgoals each tactic command-line generates.
\end{itemize}

When the correlation of these few parameters is taken within a few proof-steps, the arising statistic reveals patterns that can tell a lot about the ``meta'' proof strategy expressed by the tactics and subgoals. Coq proofs have different lengths and one small proof may potentially resemble a fragment of a bigger proof; also,
various small ``patches'' of different big proofs may resemble. We address this issue by implementing an automatic split of each proof into proof-patches, 
thus allowing ML4PG to analyse a proof by the properties of the patches that constitute the proof. The details and discussion of this feature-extraction method can be found in \cite{KHG13,HK14}.
We will not focus on the technical details of the ML4PG feature extraction here, but rather concentrate in the coming sections on proving the point that these simple 
statistical parameters (40 for one proof patch of 
five possibly composite proof steps) can indeed 
capture some essential proof-strategies, interesting and helpful enough from the user's perspective. 
%using tables where the rows represent consecutive goals of the proof 
%and colums captures some low-level features common to every goal (e.g. the three top most symbols of the goal, the tactic that is applied to
%the goal or the number of subgoals that are generated after the application of a tactic). These two dimensional arrays allow us to capture not only 
%various traceable properties of a single goal, but also transformations of the different goals throughout several proof steps. 

Once all features are extracted, ML4PG is ready to communicate with machine learning interfaces (Stage \textbf{F.2}). 
ML4PG is built to be modular -- that is, the feature
extraction is first completed within the Proof General environment, where the data is gathered in the format of hash tables, and then these tables are 
converted to the format of the chosen machine-learning tool. In~\cite{KHG13}, we connected ML4PG to several machine-learning algorithms available in Matlab~\cite{Matlab} and
Weka~\cite{Weka}; the results that we obtained with both systems were similar and Weka has the advantage of being an open-source software; hence, we  
use only Weka throughout this paper, but see \cite{KHG13} for a discussion of Matlab facilities.

ML4PG offers a choice of pattern-recognition algorithms. ML4PG is connected only to clustering algorithms~\cite{Bishop} --
a family of unsupervised learning methods. Unsupervised learning is chosen when no user guidance or class 
tags are given to the algorithm in advance: in our case, we do not expect the user to ``tag'' the library proofs in any
way. Clustering techniques divide data into $n$ groups of similar objects 
(called clusters), where the value of $n$ is provided by the user.  There are several clustering algorithms available in Weka (K-means, FarthestFirst and Expectation Maximisation, in short E.M.) 
and the user can select the algorithm using the ML4PG menu included in the Proof General interface. We show the effect of changing clustering algorithms in Table~\ref{tab:jvm}.

As will be illustrated in the later sections, various numbers of clusters can be useful: this may depend on the size of the Coq library, and on existing similarities between the proofs.
ML4PG has its own algorithm that determines the optimal number of clusters interactively, and based on the library size. As a result, 
the user does not provide the value of $n$ directly, but just decides on \emph{granularity} in the ML4PG menu, by selecting a value
between $1$ and $5$, where $1$ stands for a low granularity (producing fewer large clusters) and $5$ stands for a high granularity 
(producing many smaller clusters). Given a granularity value $g$, the number of clusters $n$ is given by the formula

$$n=\lfloor\frac{\text{objects to cluster}}{10-g}\rfloor.$$

It is worth mentioning that it is the nature of statistical methods to produce results with some probability, and not being able to provide guarantees
that a certain cluster will be found for a certain library. However, ML4PG ensures quality of the output in several different ways (Stage \textbf{F.3}).
Results of one run of a clustering algorithm may differ from another, even on the same data set. This is due to the fact that clustering algorithms randomly
choose examples to start from, and form clusters relative to those examples. However, it may happen that certain clusters are found repeatedly -- and frequently --
in different runs; then, we can use these frequencies to determine the reliable clusters.
%First of all, the results are not taken from one random run of a clustering algorithm -- instead,
In particular, ML4PG output shows a digest of clustering results coming from $200$ runs of the clustering algorithm.
The 200 runs were experimentally found to be optimal for noticing important statistics in ML4PG setting.
Only clusters that appear frequently enough are displayed to the user. 
There is a way to manipulate the frequency threshold within ML4PG, see~\cite{KHG13}. Another measure is a \emph{proximity value} assigned by clustering algorithms to every term in a cluster -- the value ranges 
from $0$ to $1$, and indicates the certainty of the given example belonging to the cluster. This proximity value is also taken into account by ML4PG before the results
are shown. If a lemma is contained in several clusters, proximity and frequency values are used to determine one ``most reliable'' cluster to display. 

ML4PG output has been recently enhanced with two new features~\cite{HK14}. In addition to the families of similar proofs, ML4PG shows the key features whose correlation determined the cluster. %, that were taken into 
%account during the cluster formation. 
Additionally, ML4PG uses this information to generate an automaton-shape representation for discovered proof-patterns and the correlated features, see Figures~\ref{automaton-nash}--\ref{automaton-nash2},~\ref{automaton-jvm1}--\ref{automaton-jvm3}.% that determined
%those patterns. This facilitates the interpretation of clusters of similar proofs.

We refer to the ML4PG user manual \cite{HK12} for a more detailed description of how to use the tool.

%\section{User scenarios}\label{sec:scenarios}

\section{User Scenario 1. Detecting Patterns in Early-Stages of the Development}\label{sec:preproof}

Users of ITPs usually start their developments loading some libraries. Those libraries contain definitions, lemmas and  theorems  
that will be used as background theory during the proof process. Some of those libraries are specific for concrete theories,
but others are common for almost every development. The common libraries contain strategies and definitions that can be extrapolated 
to other contexts; however, detecting  lemmas that follow a concrete proof-strategy %or finding a concrete definition 
can be a challenge. In this first scenario, 
we study the patterns that appear in the SSReflect library~\cite{SSReflect}.

The second purpose of this section is to set terminology and the general style of statistical proof-pattern analysis we will use throughout other sections. 
%The lemmas we show here are rather standard, the latter sections will introduce more challenging examples.

The SSReflect library extends the Coq proof language and consists of 7 files containing basic theories 
about: natural numbers, lists, booleans, functions, finite types, choice types and types with a decidable equality. 
The library contains a total of 1404 theorems; therefore, a manual inspection of these theorems to detect patterns is unfeasible. In our first scenario,
we test how ML4PG can be used to detect patterns in the SSReflect library. 

We analyse clusters that are produced in the SSReflect library using the K-means algorithm and the value $5$ as granularity parameter,
these options produce the best results in our experiments. ML4PG discovers 280 clusters using those parameters. In $45\%$ of those clusters (126 clusters), 
all the lemmas belong to the same library.  We call a cluster \emph{homogeneous} if it contains lemmas and theorems from one library, and 
\emph{heterogeneous} if it contain objects from different libraries. 

The mean size of the homogeneous clusters are 4 elements, and the similarities of the lemmas of a cluster can be easily spotted in most of the cases. From the 126 clusters, 
we can distinguish the following classification of clusters (see also~\cite{HK12} for a supporting extended note about this experiment).

\begin{itemize}
 \item $36\%$ of the clusters consist of lemmas about related functions.
 
 \begin{example}\label{example1}
Examples of this kind of clusters are the ones including lemmas about:
% 
% \noindent - \lstinline?max? and \lstinline?min? functions (\lstinline?ssrnat? library),
% for instance the cluster containing the two following lemmas:
% 
% \begin{lstlisting}
% Lemma maxn_mulr : right_distributive muln maxn.
% Lemma minn_mulr : right_distributive muln minn.
% \end{lstlisting}
% 
% \noindent - \lstinline?and? and \lstinline?or? (\lstinline?ssrbool? library), e.g. the one that consists of the lemmas:
% 
% \begin{lstlisting}
% Lemma andbb : idempotent andb.
% Lemma orbb : idempotent orb.
% \end{lstlisting}
% 
% \noindent - 
\lstinline?take? and \lstinline?drop? (\lstinline?take? takes the first $n$ elements of a list and \lstinline?drop? removes the first $n$ elements of the list):

\begin{lstlisting}
Lemma map_take s : map (take n0 s) = take n0 (map s).
Lemma map_drop s : map (drop n0 s) = drop n0 (map s).
\end{lstlisting}

 \end{example}

 \item $20\%$ of clusters contain lemmas that follow the same proof structure and that share some common auxiliary results.
 
\begin{example}\label{example2}
 Examples of this kind of cluster appears in several libraries, for instance in the \lstinline?seq? library:

\begin{lstlisting}
Lemma has_map a s : has a (map s) = has (preim f a) s.
Proof. by elim: s => //= x s ->. Qed.

Lemma all_map a s : all a (map s) = all (preim f a) s.
Proof. by elim: s => //= x s ->. Qed.

Lemma count_map a s : count a (map s) = count (preim f a) s.
Proof. by elim: s => //= x s ->. Qed.
\end{lstlisting}

% \noindent and also for the \lstinline?nat? library:
% 
% \begin{lstlisting}
% Lemma addnCA : left_commutative addn.
% Proof.by move=> m n p; elim: m => //= m; rewrite addnS => <-.
% Qed.
% 
% Lemma mulnDl : left_distributive muln addn.
% Proof. 
% by move=> m1 m2 n; elim: m1 => //= m1 IHm; rewrite -addnA -IHm.
% Qed.
% \end{lstlisting}
\end{example}
 
 \item $13\%$ of clusters consist of theorems that are used in the proofs of other theorems of the same cluster.

\begin{example}\label{example3}
ML4PG discovers that the following two lemmas are in the same cluster:

\begin{lstlisting}
Lemma altP : alt_spec b.
Lemma boolP : alt_spec b1 b1 b1. Proof. exact: (altP idP). Qed. 
\end{lstlisting}
\end{example}

 \item $11\%$ of clusters are formed by ``view'' lemmas, an important kind of lemmas that are used in SSReflect to apply boolean reflection~\cite{SSReflect}. 
 
\begin{example}\label{example4}
ML4PG finds a cluster with the following two view lemmas coming from the \lstinline?fintype? library:

\begin{lstlisting}
Lemma unit_enumP : Finite.axiom [::tt]. Proof. by case. Qed.
Lemma bool_enumP : Finite.axiom [:: true; false].
Proof. by case. Qed.
\end{lstlisting} 
\end{example}

 \item $5\%$ of the clusters contain equivalence lemmas that are proven just by simplification. 
 
\begin{example}
 An example of this kind of clusters is given by the cluster that contains the following lemmas:
 
\begin{lstlisting}
Lemma multE : mult = muln.     Proof. by []. Qed.
Lemma mulnE : muln = muln_rec. Proof. by []. Qed.
Lemma addnE : addn = addn_rec. Proof. by []. Qed.
Lemma plusE : plus = addn. Proof. by []. Qed.
\end{lstlisting}
 
\end{example}
 
 \item $4\%$ of the clusters consist of lemmas that are solved using analogous lemmas. 
 
 \begin{example}\label{example5}
 
An example of clusters that consists of lemmas that are solved using analogous lemmas is the one containing
the following two lemmas. 
 
\begin{lstlisting}
Lemma addnAC : right_commutative addn.
Proof. by move=> m n p; rewrite -!addnA (addnC n). Qed. 
 
Lemma subnAC : right_commutative subn.
Proof. by move=> m n p; rewrite -!subnDA addnC. Qed. 
\end{lstlisting}   

\noindent Namely, the lemma \lstinline?subnDA? (\lstinline?forall (a b c : nat), a-(b+c) = (a-b)-c?) could be obtained from lemma 
\lstinline?addnA? (\lstinline?forall (a b c : nat), a+(b+c) = a+b+c?) using techniques like lemma analogy~\cite{lpar13}.
 \end{example}

\end{itemize}

% % From these 126 clusters, the : for instance, ML4PG discover cluster contains lemmas about \lstinline?leq? ($<=$) and 
% % \lstinline?lgt? ($<$), \lstinline?max? and \lstinline?min? functions, \lstinline?and? and \lstinline?or? operators, and \lstinline?take? and \lstinline?drop? list operators
% % (\lstinline?take? takes the first $n$ elements of a list and \lstinline?drop? removes the first $n$ elements of the list).
% 
% The $20\%$ of clusters have lemmas that follow the 
% same proof structure and that share some common auxiliary results, and a $13\%$ of clusters consists of theorems that are used in the proofs of other theorems of the 
% same cluster. The $11\%$ of clusters are view lemmas, an important kind of lemmas that are used in SSReflect to apply boolean reflection~\cite{SSReflect}. 
% A $5\%$ of the lemmas are equivalence lemmas that are proven just by simplification. 

In the case of heterogeneous clusters (clusters that include lemmas from different libraries), ML4PG discovers 154 clusters. In this case, the size of the clusters is
bigger than in the case of homogeneous clusters; namely, the mean size is 8 lemmas per cluster. The different clusters can be classified as follows.

\begin{itemize}
 \item $31\%$ of the clusters contain lemmas that state properties applicable to several operators from different libraries.
 
 \begin{example}\label{example6}

 ML4PG discovers a cluster containing lemmas about the associativity of the addition of natural numbers (\lstinline?addn? function) and the 
 associativity of the concatenation of lists (\lstinline?++? operator).
 
\begin{lstlisting}
Lemma catA s1 s2 s3 : s1 ++ s2 ++ s3 = (s1 ++ s2) ++ s3.
Proof. by elim: s1 => //= x s1 ->. Qed. 
 
Lemma addnA : associative addn.
Proof. by move=> m n p; rewrite (addnC n) addnCA addnC. Qed. 
\end{lstlisting}
  
 \end{example}

 \item $27\%$ of the clusters consist of lemmas related to operations over the base case of types.
 
\begin{example}\label{example7}

As an example of this kind of clusters, ML4PG discovers that there is a strong correlation among the following four lemmas:
 
 \begin{lstlisting}
Lemma andTb : left_id true andb. Proof. by []. Qed.
Lemma orFb : left_id false orb. Proof. by []. Qed.
Lemma mul0n : left_zero 0 muln. Proof. by []. Qed. 
Lemma sub0n : left_zero 0 subn.    Proof. by []. Qed.
\end{lstlisting}
 
\end{example}

 \item $12\%$ of the clusters come from lemmas whose proof rely on the fundamental lemmas.
 
 \begin{example}\label{example8}
  
 ML4PG discovers a cluster with the following two lemmas about \lstinline?rot?  (that rotates a list \lstinline?l? left \lstinline?n? times) and the \lstinline?expn? (exponentiation 
 function). 
 
 \begin{lstlisting}
Lemma rot0 s : rot 0 s = s.
Proof. by rewrite /rot drop0 take0 cats0. Qed. 

Lemma expn_eq0 m e : (m ^ e == 0) = (m == 0) && (e > 0).
Proof. by rewrite !eqn0Ngt expn_gt0 negb_or -lt0n. Qed.
\end{lstlisting}

\noindent At first sight, it seems that the only similarity between these two lemmas is that they only use rewriting rules in their proofs, however if we carefully inspect the lemmas
that are used for rewriting, we notice that most of them are fundamental lemmas about \lstinline?nil? (the base constructor for the \lstinline?list? type)
and \lstinline?0? (the base constructor for the \lstinline?nat? type).

 \end{example}

 \item $9\%$ of the clusters combine lemmas from the libraries about lists and natural numbers -- note that the definition of lists
and natural numbers is quite similar, both have one base case and a recursive one, so several lemmas are solved applying induction and using the inductive hypothesis.  

\begin{example}\label{example9}
 
 An example of this kind of clusters is given by the one that consists of the following lemmas:
 
\begin{lstlisting}
Lemma catrev_catr s t u : catrev s (t ++ u) = catrev s t ++ u.
Proof. by elim: s t => //= x s IHs t; rewrite -IHs. Qed.

Lemma mulnDl : left_distributive muln addn.
Proof. 
  by move=> m1 m2 n; elim: m1 => //= m1 IHm;
  rewrite -addnA -IHm.
Qed.

Lemma mem_cat x s1 s2: 
    (x \in s1 ++ s2) = (x \in s1) || (x \in s2).
Proof. by elim: s1 => //= y s1 IHs; rewrite !inE /= -orbA -IHs.
Qed. 
\end{lstlisting}

\noindent In all these lemmas, we can see that induction is applied and after the use of some rewriting rules the inductive hypothesis is applied to finish the proof.

\end{example}

\end{itemize}

% 
% $31\%$ of these clusters are lemmas that state properties that can be applied to several operators from different libraries 
% (e.g. associativity for \lstinline?cat? operator -- concatenation of lists -- and \lstinline?add? function, or inner commutativity of \lstinline?or? operator and \lstinline?add?
% function), the proof of all these lemmas is similar but it is adapted to the concrete library. The $27\%$ of clusters consists of lemmas related to operations over 
% base case of types (e.g. the lemmas \lstinline?andTb : forall x, true && x = true? and \lstinline?mul0n: forall x, 0 * n = 0? are in the same cluster), these lemmas can be 
% easily proved applying cases and simplification. The lemmas of the previous clusters are fundamental lemmas, the similarity of a $12\%$ of the clusters come from lemmas whose
% proof rely on the fundamental lemmas. A $9\%$ of the clusters combine lemmas from the libraries about lists and natural numbers -- note that the definition of lists
% and natural numbers is quite similar, both have one base case and a recursive one, so several lemmas are solved applying induction and using the inductive hypothesis. 

The similarity of most clusters ($83\%$ of them) can be easily discovered just inspecting the statement of the lemmas and their proofs. However, clustering is a statistical 
tool and 
there can be marginal cases, when one of the following situations arise:

\begin{enumerate}
	\item[$S1$] The correlation of proof features is weak, e.g. it is a ``leftover'' cluster containing proofs that were not grouped with other strongly correlating clusters. 
	In most of those cases, the clusters contain more than 10 elements, %and we can discover
%patterns among subsets of those clusters, 
but it is difficult to find a common pattern followed by all lemmas.
		\item[$S2$] There may be a strong feature correlation within a cluster, but the user does not understand what it is. 
		
\noindent For $S1$-$S2$, we suggest to use automaton-shape proof-analysis of ML4PG; as we explain in the next sections.
The basic idea is that the automaton shows whether there was a strong feature correlation, and if yes, -- what it was; so that the user can identify in which of the situations he currently  is.
		
		\item[$S3$] The user may see the correlated features in the ML4PG automaton, but  nevertheless does not find this a useful suggestion.
For case $S3$, there is little ML4PG can do but help the user to understand that his case is exactly $S3$ (in which case, there is  a strong correlation of $S2$, but it is irrelevant). 
One could try modifying various ML4PG parameters (like granularity, clustering algorithm, proof library), in a search for a better cluster.
But ultimately, the user's ``desired'' proof may simply be different from any previous proof; so there may not be a better pattern in the library.

\end{enumerate}

%in some cases there is not a clear correlation among the lemmas of a cluster. 

The above results show that ML4PG can be useful to detect patterns in early stages of a development. Namely, it can be used to find relations among
functions and their lemmas, common strategies followed in a library and fundamental lemmas applied in several proofs. Besides, if a user knows a library (e.g.
the library defining natural numbers), ML4PG can show similarities 
between lemmas about natural numbers and lists, facilitating the use of the new library based on the previous knowledge of the user.

\section{User Scenario 2. ML4PG for Detecting Irrelevant Libraries}\label{sec:nash}

An (abstract) sequential game can be represented as a tree with pay-off functions in the leaves, dictating the win or loss of each player when the game finishes there.
Each internal node is owned by a player and a play of a game is a path from the root to a leaf. A \emph{strategy} is a game where each internal node has 
chosen a child. A \emph{Nash equilibrium} is a strategy in which no agent can change one or more of his choices to obtain a better overall result for himself.
A strategy is a \emph{subgame perfect equilibrium} if it represents/have a Nash equilibrium of every subgame of the original game. 

In this scenario, we use ML4PG to analyse two Coq libraries that formalise that all sequential games have Nash equilibria in binary games
(games where each internal node has two choices)~\cite{Ves06}  and in the general case~\cite{nash}. 
Note that unlike the other benchmarks presented throughout the paper,
the files presented here are developed using plain Coq instead of SSReflect.
ML4PG adapts to this change automatically.
% This does not mean any hindrance to ML4PG that can work with both plain
%Coq and its SSReflect dialect. 

\begin{figure}[t]
\centering

 	\scriptsize{
 		\begin{tabular}{|l|l|}
 		\hline
 	\lstinline?BI_Exists? & \lstinline?NashEq_Exists? \\
 		\hline
 		\hline
                {\scriptsize \lstinline?Theorem BI_Exists :?} & {\scriptsize \lstinline?Theorem NashEq_Exists :?}\\
 		{\scriptsize ~~~~\lstinline?forall g, exists s, BI s /\ g = s2g s.?} & {\scriptsize ~~~~\lstinline?forall g, exists s, NashEq s /\ g = s2g s.?}\\
 		{\scriptsize \lstinline?Proof. deskolem_apply BI_fctExists. Qed.?} & {\scriptsize  \lstinline?Proof. deskolem_apply NashEq_fctExists. Qed.?}\\
 		&\\
 		{\scriptsize \lstinline?Theorem BI_fctExists : exists F, forall g,?} &  {\scriptsize \lstinline?Theorem NashEq_fctExists : exists F, forall g,?}\\
                {\scriptsize ~~~~\lstinline? BI (F g) /\ g = s2g (F g).?} & {\scriptsize ~~~~\lstinline? NashEq (F g) /\ g = s2g (F g).?}\\
                {\scriptsize \lstinline?Proof. ?} & {\scriptsize \lstinline?Proof. ?}\\
                {\scriptsize \lstinline?exists compBI. intro g. split.?} & {\scriptsize \lstinline?exists compBI. intro g. split.?}\\
                {\scriptsize ~~~~~~~~\lstinline? exact (compBI_is_BI g).?} & 
                        {\scriptsize ~~~~~~~~\lstinline? apply BI_is_NashEq. exact (compBI_is_BI g).?}\\
                {\scriptsize \lstinline? exact (s2g_inv_compBI g).?} & 
                        {\scriptsize \lstinline? exact (s2g_inv_compBI g).?}\\
               {\scriptsize \lstinline?Qed.?} & 
                        {\scriptsize \lstinline?Qed.?}\\        
 		\hline
 		\end{tabular}
 		
 		}

\includegraphics[scale=.28]{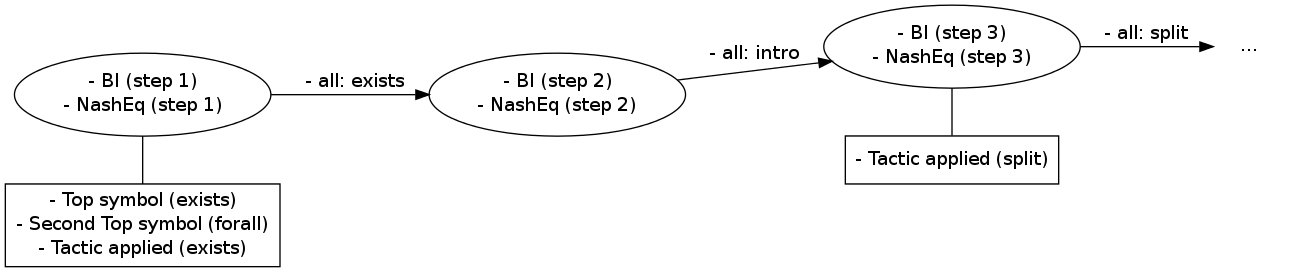}
\caption{{\scriptsize\textbf{Top. Proofs of theorems \texttt{BI\_Exists} and \texttt{NashEq\_Exists}, \emph{coming from one library~\cite{Ves06}; and grouped together by ML4PG. Bottom. Fragment of the automaton-shape representation of the cluster containing the theorems.}}
\emph{Note that there is always just one transition between the neighbouring states, which suggests that similar or same tactics are applied. According to the picture, the features with the highest correlation are the first two top symbols in the first goal; and the first and third applied tactics. Note that the second tactic -- \texttt{intro} is not mentioned among the correlating features, as it is a common second-step tactic in many other proofs, outside of the cluster.}}}\label{automaton-nash} 
\end{figure}

\begin{figure}[t]
\centering
 	\scriptsize{
 		\begin{tabular}{|l|l|}
 		\hline
 	Binary case & General case \\
 		\hline
 		\hline
                {\scriptsize \lstinline?Lemma SGP_is_NashEq : ?} & {\scriptsize \lstinline?Lemma SPE_is_Eq :  ?}\\               
                {\scriptsize \lstinline?forall s : Strategy, SGP s -> NashEq s.?} & {\scriptsize \lstinline?forall s : Strat, SPE s -> Eq s. ?}\\               
                {\scriptsize \lstinline?Proof.?} & {\scriptsize \lstinline?Proof.?}\\       
                {\scriptsize \lstinline?induction s.?} & {\scriptsize \lstinline?intros. destruct s. simpl in H. tauto. ?}\\                      
                {\scriptsize \lstinline?unfold NashEq. intros _.  induction s'.?} & {\scriptsize \lstinline?Qed.?}\\                      
                {\scriptsize \lstinline?intros. unfold stratPO. unfold agentConv in H. ?} &  \\                      
                {\scriptsize \lstinline?rewrite (H a). trivial. ?} &  \\   
                {\scriptsize \lstinline?unfold agentConv. intros. contradiction.?} &  \\    
                {\scriptsize \lstinline?unfold SGP. intros [_ [_ done]]. trivial.?} &  \\    
 		{\scriptsize \lstinline?Qed.?} &  \\        
 		\hline
 		\end{tabular}
 		
 		}
\includegraphics[scale=.23]{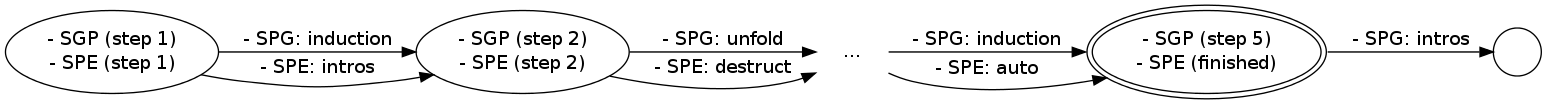}
\caption{{\scriptsize{\textbf{Top. Proofs of the theorems stating that Subgame Perfect Equilibrium implies Nash Equilibrium.}
 	\emph{The lemma statements are very similar; however, the structure of the proofs is completely different; hence, ML4PG does not group these proofs together.}
	\textbf{Bottom. Fragment of automaton-shape representation of the proofs of the theorems.} 
	\emph{In contrast to the automaton of Figure~\ref{automaton-nash}, each proof step gives a separate transition (with a different tactic) and the proof-goals do not contain any features that correlate.}}}}\label{automaton-nash2} 
\end{figure}

%We provide a similar analysis to the one introduced in Section~\ref{sec:preproof} about the SSReflect library. In this case, 
It would be natural to assume that the proofs involved in verification of the two results will be very similar, and thus one could potentially hope for proof-pattern re-use. 
However, close inspection of these libraries can reveal that the actual proof strategies used in both libraries are different.
Without ML4PG, such ``negative'' discovery would require user time and experience in comparing Coq proofs.
We instead give it as a challenge to ML4PG that takes only a few seconds to analyse the libraries.
ML4PG loads the Coq files 
developed in~\cite{Ves06,nash} and a library about topological sorting~\cite{ler07}  used in~\cite{nash}. These Coq files include 145 theorems,
and we choose the K-means algorithm and the value $5$ as the  granularity parameter to obtain clusters using ML4PG. ML4PG finds 32 clusters using those parameters, and 
their mean size is three elements per cluster. The question is:  how can the user interpret these results, %these clusters, 
when he sees those 32 sets of approximately three lemmas/theorems on the Proof General screen? One way to understand the clusters is to analyse their relative statistics and the structure of proofs contained in the clusters, 
as we explain below. Another option that ML4PG provides is visualising the common proof patterns, as e.g. shown in Figures~\ref{automaton-nash}--\ref{automaton-nash2},~\ref{automaton-jvm1}--\ref{automaton-jvm3}.

The automata produced by ML4PG should be read as follows. The automaton represents the proof-patches (5 proof-steps) of the lemmas that were clustered together. 
  There are at most $5$ states in the automaton and the $i$th state is given by the list of $i$th proof-steps of the proof-patches. The transitions between the states represent 
  the tactics applied in the $i$th proof-steps.  If the tactics of several $i$th proof-steps are the same or are related, the transitions are merged -- and the automaton shows only one transition, cf.~Figure~\ref{automaton-nash}.
	If, on the contrary, the tactics are different, there will be several different transitions shown between the states; annotated with the different tactics.
	Finally, if features of any particular proof step played a special role in the process of associating the proof patches in the cluster, the correlated features annotate that proof-state and are shown in a square box (thus the box itself is not a state or part of the automaton).

%\begin{table}[t]
% 	\centering

 %	\caption{{\scriptsize\textbf{Proofs of theorems \texttt{BI\_Exists} and \texttt{NashEq\_Exists},} \emph{coming from one library~\cite{Ves06}; and grouped together by ML4PG.}}}
 %	\label{tab:nash0}
 %\end{table}

It can be easily seen from ML4PG annotation of results that 21 of the 32 clusters ($65\%$) in the Nash libraries are 
homogeneous clusters, thus the similarity between the proofs within one library is higher than across libraries. 
Starting first with those homogeneous clusters, we notice that 
%obtain a classification of clusters that is similar to the one presented in
%Section~\ref{sec:preproof}.

%\begin{itemize}
 $\bullet$ 8 clusters ($38\%$) contain lemmas about related functions, -- as they use similar lemmas in their proofs.  
 
 \begin{example}[\textbf{Correlating homogeneous clusters in the Nash libraries -- $S2$ scenario}]
  As an example of this kind of clusters, ML4PG discovers a cluster with two lemmas from~\cite{Ves06}: the first one (\lstinline?BI_Exists?) states that for every game, 
  there exists a strategy that makes the game to have Backward-Induction equilibrium (each player plays optimally at every node); the second lemma (\lstinline?NashEq_Exists?)
  states the analogous result for the Nash equilibrium. 
    From the proofs of these two lemmas, it is clear that they use a similar sequence of tactics. 
	This is also reflected in the automaton generated by ML4PG, see Figure~\ref{automaton-nash}. 
	We know it is Scenario $S2$, rather than $S1$, as all transitions are single arrows, and the correlated features are shown by the automaton.
  %state is annotated with the list of features that determined the proof-pattern. 
 \end{example}

% \item 
$\bullet$ 6 clusters ($28\%$) consist of lemmas about a concrete function.
 
  \begin{example}
  In~\cite{nash}, there is a function called \lstinline?StratPref? that given an agent and two strategies decides which is the best one. 
  ML4PG finds a cluster with two lemmas: the first one (\lstinline?StratPref_dec?) states the decidability of the function; and the second one states that the function 
  produces an irreflexive relation. 
 \end{example}
 
% \item 
$\bullet$ 4 clusters ($19\%$) contain theorems that use other theorems of the cluster in their proofs. 
 
%\end{itemize}

%\begin{table}[t]
 %	\centering

 	%\caption{{\scriptsize\textbf{Proof of the theorem stating that Subgame Perfect Equilibrium implies Nash Equilibrium. Left.} \emph{Binary case}. \textbf{Right.} \emph{General case.} 
 	%\emph{The lemma statements are very similar; however, the structure of the proofs is completely different; hence, ML4PG does not group these proofs together. } }}
 	%\label{tab:nash}
 %\end{table}
 
This quick analysis would show that some obvious grouping of proofs within one library was made by ML4PG. But, unless we are interested in a particular proof technique 
appearing in one of them, we direct our attention to patterns found across the libraries, hoping to find some common proof methods across the developments.

In the case of heterogeneous clusters, all the clusters consist of lemmas about auxiliary functions (for instance, about different properties of lists) that are common in all 
the libraries we are studying. However, there is no correlation among the important theorems of these libraries.

\begin{example}[\textbf{Disagreeing heterogeneous proofs in the Nash library}]
The proofs of the two main theorems of the two Nash libraries are given in Figure~\ref{automaton-nash2}. Even if~\cite{nash} is a generalisation of the work presented
in~\cite{Ves06}, the proofs for Nash equilibrium are completely different, mainly for two reasons. First, the datastructures that are used in each development are too different, 
and therefore the lemmas about them do not have a strong correlation. In addition, the approaches that are used to prove them are completely different: one based on a procedure called backward induction~\cite{Ves06} and the other is based on the fact that the preference of players is acyclic. 
\end{example}

	The above two proofs were not suggested in one cluster by ML4PG. However, if they were two ``leftover'' proofs, as Scenario $S1$ explains, their automaton would look as Figure~\ref{automaton-nash2} suggests. 
	The user would determine that it is an $S1$ (rather than $S2$) case, by noticing that
	%This can be noticed by inspecting the proofs directly or by looking at the automaton of .
		%		We know it is the scenario $S1$, rather than $S2$, as 
		transitions between every two neighbouring states are different, and no correlated features are shown by the automaton.

The results do not vary much when we try changing the clustering algorithm and the granularity values -- reducing the granularity value produces bigger homogeneous clusters, but has little effect on heterogeneous clusters. 
As can be seen from this example,
the  ML4PG-based proof-pattern check could be an easy and fast way of getting the information about the
absence of recyclable patterns across the libraries.

Note that  in this case study, the total number of lemmas is smaller than in the 
previous case-study (145 theorems); 
but the feature extraction mechanism of ML4PG automatically adapts to this, and handles 
statistics of small data sets as well as statistics of bigger data sets.
%however, this does not mean any hindrance to ML4PG whose performance is not affected by the size of the datasets.  

%\input{scenarios}

\section{User Scenario 3. A Team-Based Development}\label{sec:JVM}

In the last scenario, we turn to team-based applications of Coq and ML4PG. For this purpose,
we translate the ACL2 proofs of correctness of the Java Virtual Machine (JVM)~\cite{M03} into Coq/SSReflect. 
JVM~\cite{JVM} is a stack-based abstract machine which can execute Java bytecode. We have modelled an interpreter for
JVM programs in Coq/SSReflect. From now on, we refer to our machine as ``SJVM'' (for SSReflect JVM).

\begin{figure}
\begin{minipage}{0.3\linewidth}
\centering
{\scriptsize
\begin{lstlisting}
static int factorial(int n)
{
  int a = 1;
  while (n != 0){
    a = a * n;
    n = n-1;
    }
  return a;
}
\end{lstlisting}}
\end{minipage}
\begin{minipage}{0.175\linewidth}
\centering
{\scriptsize
$\begin{array}{ccl}
0&:& iconst~1\\
1&:& istore~1\\
2&:& iload~0\\
3&:& ifeq~13\\
4&:& iload~1\\
5&:& iload~0\\
6&:& imul\\
7&:& istore~1\\
8&:& iload~0\\
9&:& iconst~1\\
10&:& isub\\
11&:& istore~0\\
12&:& goto~2\\
13&:& iload~1\\
14&:& ireturn\\
\end{array}$}
\end{minipage}
\hspace{0.5cm}
\begin{minipage}{0.35\linewidth}
\centering
{\scriptsize
\begin{lstlisting}
Fixpoint helper_fact (n a) :=
match n with 
| 0 => a
| S p => helper_fact p (n * a)
end.

Definition fn_fact (n : nat) := helper_fact n 1.
\end{lstlisting}}
\end{minipage}

\caption{{\scriptsize\emph{\textbf{Factorial function}. \textbf{Left:} Java program for computing the factorial of natural numbers. \textbf{Centre:} Java bytecode
associated with the Java program. \textbf{Right:} tail recursive version of the factorial function in Coq/SSReflect.}}}\label{fig:factorial_bytecode} 
 
\end{figure}

An industrial scenario of interactive theorem proving may involve distribution of work-load across a team, and a bigger proportion of 
routine or repetitive cases. Here, the inefficiency often arises when programmers use different notation to accomplish very similar tasks,
and thus a lot of work gets duplicated, see also~\cite{BHJM12}. We tested ML4PG in exactly such a scenario: we assumed that a programming 
team is collectively developing proofs of \emph{the soundness of the specification, and the correctness of the implementation}
of Java bytecode for a dozen of programs computing multiplication, powers, exponentiation, and other functions about natural numbers. 
A new team member then tries to learn the important proof patterns while trying to prove %the soundness of specification, and correctness of implementation of a 
similar results for a new function -- factorial.

Given a specific Java method, we can translate it to Java bytecode using a tool such as
\lstinline?javac? of Sun Microsystems. Such a bytecode can be executed in SJVM 
provided a schedule (a list of thread identifiers indicating the order in which the threads are to be stepped), and the result will be the state of the JVM at the end of the schedule. 
Moreover, we can prove theorems about the SJVM model behaviour when interpreting that 
bytecode. 

\begin{example}
The bytecode associated with the factorial program can be seen
in Figure~\ref{fig:factorial_bytecode}.
\end{example}

The state of the SJVM consists of 4 fields: a \emph{program counter} (a natural number), a set of registers called \emph{locals}
(implemented as a list of natural numbers), an operand \emph{stack} (a list of natural numbers), and the bytecode \emph{program}
of the method being evaluated.

Java bytecode, like the one presented in Figure~\ref{fig:factorial_bytecode}, can be executed within SJVM. However, more interestingly than merely executing
Java bytecode, we can prove the correctness of the implementation of the Java bytecode programs using Coq/SSReflect. 
For instance, in the case of the factorial program, the new team member is asked to prove the following theorem, which states the correctness of the \lstinline?factorial? bytecode. 

\begin{thm}\label{lem:factorial}
Given a natural number $n$ and the factorial program with $n$ as an input, SJVM produces a state which contains $n!$ on top of the stack
 running the bytecode associated with the program.  
\end{thm}

%\begin{table}
% 	\centering
% 	\scriptsize{

% 		}
% 	\caption{{\scriptsize\textbf{Proofs of equivalence of the tail-recursive and 
% 	recursive versions of functions exponentiation and factorial, following Proof Strategy~\ref{ps:spec}.} \emph{The left-hand-side shows 
%	a few initial proof steps for \texttt{fn\_fact\_is\_theta}, leading to a deadlock. The right-hand-side shows the  lemma (\texttt{fn\_expt\_is\_theta}) suggested by ML4PG  (see Table~\ref{tab:jvm}) and an auxiliary lemma used to prove it. In italics is the proof reconstruction by analogy.}}}
% 	\label{tab:jvmproofs}
% \end{table}

The proof of theorems like the one above  always follows the same methodology adapted from ACL2 proofs about Java Virtual Machines~\cite{M03}
and which consists of the following three steps.

\begin{enumerate}
 \item[(1)] Write in Coq/SSReflect the specification of the function and the algorithm, and prove that the algorithm satisfies the specification.
 \item[(2)] Write the JVM program within Coq/SSReflect, define the function that schedules the program (this function will make SJVM
 run the program to completion as a function of the input to the program), and prove that the resulting code implements this algorithm.
 \item[(3)] Prove total correctness of the Java bytecode.
\end{enumerate}

Using this methodology, we have proven the correctness of several programs related to arithmetic (multiplication of natural
numbers, exponentiation of natural numbers, and so on); see \cite{HK12}. The proof of each theorem was done independently from others to  model a distributed proof development.

Therefore, we simulate the following scenario.
Suppose a new developer tackles for the first time the proof of Theorem~\ref{lem:factorial}, and he knows the
general methodology to prove it and has access to the library of programs previously proven by other users. In this situation 
the different notations employed by different users obscure some common features.
ML4PG would be a good alternative to the manual search for proof patterns.

\begin{figure}
\centering
\scriptsize{
 		\begin{tabular}{|l|l|}
 		\hline
 	Factorial & Exponentiation \\
 		\hline
 		\hline
 {\scriptsize \lstinline?Lemma fn_fact_is_theta n : fn_fact n = n`!.? }& {\scriptsize \lstinline?Lemma fn_expt_is_theta n m : fn_expt n m = n^m.?}\\
{\scriptsize \lstinline?Proof.?} & {\scriptsize \lstinline?Proof.?}\\
{\scriptsize \lstinline?rewrite /fn_fact.? } & {\scriptsize \lstinline?by rewrite /fn_expt helper_expt_is_theta ?}\\
{\scriptsize \textit{\lstinline?by rewrite helper_fact_is_theta mul1n. ?}} & {\scriptsize ~~~~~~~~~~\lstinline?mul1n.?}\\
{\scriptsize \textit{\lstinline?Qed.?}} & {\scriptsize \lstinline?Qed.?}\\
                             & \\		
{\scriptsize \textit{\lstinline?Lemma helper_fact_is_theta n a :?}} & {\scriptsize \lstinline?Lemma helper_expt_is_theta n m a :  ?} \\
 {\scriptsize ~~~~~\textit{\lstinline? helper_fact n a = a * n`!.?}} & {\scriptsize ~~~~~\lstinline?helper_expt n m a = a * (n ^ m).?} \\
{\scriptsize \textit{\lstinline?Proof.?}} & {\scriptsize \lstinline?Proof.?}\\
{\scriptsize \textit{\lstinline?move : n a; elim : m => [a m| m IH n a /=].?}} & {\scriptsize \lstinline?move : a; elim : n => [a| n IH a /=].?}\\
{\scriptsize ~~~~~~~~~~\textit{\lstinline?  by rewrite /theta_fact fact0 muln1.?}} & {\scriptsize ~~~~~~~~~~\lstinline?  by rewrite /theta_expt expn0 muln1.?}\\
{\scriptsize \textit{\lstinline?by rewrite IH /theta_fact factS ?}} & {\scriptsize \lstinline?by rewrite IH /theta_expt expnS?}\\
{\scriptsize ~~~~~~~~~~\textit{\lstinline?mulnA [a * _]mulnC.?}} & {\scriptsize ~~~~~~~~~~\lstinline?mulnA [a * _]mulnC.  ?}\\
{\scriptsize \textit{\lstinline?Qed.?}} & {\scriptsize \lstinline?Qed.?}\\

 		\hline
 		\end{tabular}
}
\includegraphics[scale=.3]{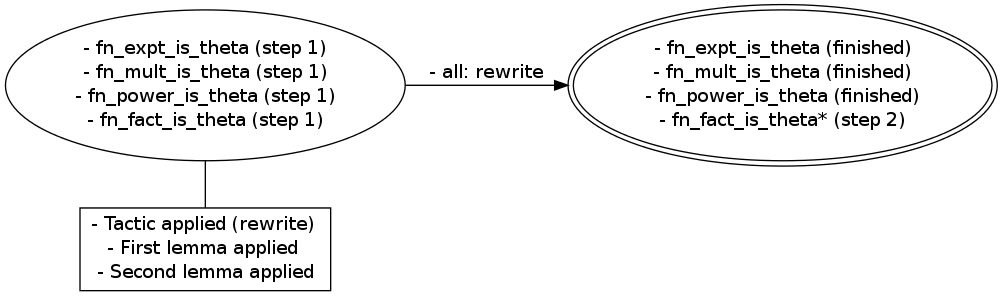}
\caption{{\scriptsize{\textbf{Top. Proofs of equivalence of the tail-recursive and 
 	recursive versions of functions exponentiation and factorial, following Proof Strategy~\ref{ps:spec}.} \emph{The left-hand-side shows 
	a few initial proof steps for \texttt{fn\_fact\_is\_theta}, leading to a deadlock. The right-hand-side shows the  lemma (\texttt{fn\_expt\_is\_theta}) suggested by ML4PG  (see Table~\ref{tab:jvm}) and an auxiliary lemma used to prove it. In italics is the proof reconstruction by analogy.} \textbf{Bottom. Automaton-shape representation of the proofs following Proof Strategy~\ref{ps:spec}.} \emph{The automaton only contains two states and its single transition 
is given by the \texttt{rewrite} tactic. Note that the three important features for this pattern are given by the applied  tactic, the first lemma applied (a lemma that expands 
the definition of \texttt{fn\_fact} and \texttt{fn\_expt}) and the second lemma applied (the auxiliary lemma about the \texttt{helper} function). 
Note that the box explicitly suggests the lemmas to formulate for the proof of \texttt{fn\_fact\_is\_theta}!
(The $*$ in the the second step of \texttt{fn\_fact\_is\_theta} indicates the last proof step of the incomplete proof.)}}}}\label{automaton-jvm1} 
\end{figure}

Let us focus on the first step of the methodology -- that is, the proof of the equivalence between the specification of the factorial function (which is already defined in
SSReflect using the function \lstinline?factorial? having the notation \lstinline?n`!? for \lstinline?factorial n?) 
and the algorithm, see Figure~\ref{automaton-jvm1}. The Java factorial function is an iterative function; and
the algorithm is written in Coq as a tail recursive function, see the right side of Figure~\ref{fig:factorial_bytecode}. In the available SJVM libraries, all
the tail recursive functions are defined using an auxiliary function, called the \emph{helper}, and a wrapper for such a function. 
Discovering this fact is the first challenge for ML4PG. Suppose the new team member has stopped after one proof-step of trying to prove the lemma \texttt{fn\_fact\_is\_theta} in a naive way, without a helper function; see Figure~\ref{automaton-jvm1}. He cannot proceed, and calls ML4PG for a hint.
The suggestions
provided by ML4PG in this case are the proofs of step $(1)$ for three iterative programs: the multiplication, the exponentiation and the power of
natural numbers. %; see e.g. Lemma \texttt{fn\_expt\_is\_theta} in Table~\ref{tab:jvmproofs}. 
In addition, ML4PG produces the automaton-shape representation of Figure~\ref{automaton-jvm1} showing 
the trace of the four proofs and the correlated features that determined the pattern. From the proofs of those lemmas, and the information in the automaton, it is easy to notice that all of them use an auxiliary lemma (like \texttt{helper\_expt\_is\_theta}), and thus follow the same proof strategy: % which can be also applied in the case of factorial: 

\begin{PS}\label{ps:spec}
 
\emph{Prove an auxiliary lemma about the helper considering the most general case. For example, if the helper function is defined with formal parameters
$n$, $m$, and $a$, and the wrapper calls the helper initializing $a$ at $0$, the helper theorem must be about \lstinline?(helper n m a)?, not just about the special
case \lstinline?(helper n m 0)?. Subsequently, instantiate the lemma for the concrete case.}
 
\end{PS}

\begin{table}
\centering
{\footnotesize
 \begin{tabular}{|c|c|c|c|c|c|c|c|c|c|c|c|c|c|c|c|}
   \hline
    & $g=1$& $g=2$& $g=3$& $g=4$ & $g=5$\\
      Algorithm: & ($n=16$) & ($n=18$) & ($n=21$) & ($n=24$) & ($n=29$)\\
   \hline
   K-means &30$^{a,b,d}$   &\cellcolor[gray]{0.9}\textbf{4}$^{a-d}$  &\cellcolor[gray]{0.9}\textbf{4}$^{a-d}$ & 2$^{c,d}$& 0 \\
   \hline
   E.M. &21$^{a-d}$  &7$^{a-d}$  & 7$^{a-d}$ &0 &0  \\
   \hline
   FarthestFirst &28$^{a-d}$ & 25$^{a-d}$&  0 &0 &0  \\
   \hline
   
  \end{tabular} }
  \caption{{\scriptsize\textbf{A series of clustering experiments discovering Proof Strategy~\ref{ps:spec}.} \emph{The table shows the sizes of clusters containing: 
  $a)$ Lemma about JVM multiplication program, $b)$ Lemma about JVM power program, $c)$ Lemma about JVM exponentiation program, and $d)$ Lemma about JVM factorial
  program. The size of the data set is 147 lemmas, in bold (grey cells) is the cluster that finds exactly the four benchmark examples. Again, the lemmas grouped by clusters are consistently 
  found for various algorithms and granularity values; and the K-means algorithm provides the most accurate clusters using $3$ as granularity value.}}}\label{tab:jvm}
\end{table}
	
%Discovery of proofs following Proof Strategy \ref{ps:spec} among  15 libraries related to formalisations of arithmetic JVM programs (a total of 147  lemmas)
%was our next benchmarking task for ML4PG, and it gave the following results. 
%\begin{example}
The technical details are as follows. ML4PG correctly suggested similar lemmas to lemma \lstinline?fn_fact_is_theta?. % in the libraries for multiplication, exponentiation and power. %, see Table~\ref{tab:jvmproofs} for the
%proofs of some of these lemmas.
%, if we use the 
%following settings:  
Table~\ref{tab:jvm} shows the results for different choices of algorithms and parameters, and we highlight  the most precise and helpful ML4PG result. In case the user is unsure of the optimal machine-learning parameters, he could use  a ``top-down approach''.
The highest granularity level does not produce any result. But, if we decrease 
the granularity level to 4, ML4PG spots some interesting similarities using the K-means algorithm. If  this is not enough to discover Proof Strategy~\ref{ps:spec}, one can decrease the granularity level 
to $3$, for which ML4PG discovers four lemmas following the same general scheme. % presented in this section. 

\begin{figure}[!]
\centering
   	\scriptsize{
 		\begin{tabular}{|l|}
 		\hline
 Factorial \\
 		\hline
 		\hline
 {\scriptsize \lstinline?Lemma program_is_fn_fact n :  ?}\\
{\scriptsize \lstinline?run (sched_fact n) (make_state 0 [::n] [::] pi_fact) =?}\\
 {\scriptsize \lstinline?  (make_state 14 [::0;fn_fact n ] (push (fn_fact n ) [::]) pi_fact).?}\\
{\scriptsize \lstinline?Proof.?}\\
 {\scriptsize \lstinline?rewrite run_app.?}\\
 {\scriptsize \textit{\lstinline?rewrite loop_is_helper_fact.?}}\\
{\scriptsize \textit{\lstinline?Qed.?}}\\
 \\
{\scriptsize \textit{\lstinline? Lemma loop_is_helper_fact n a :?}}\\
{\scriptsize \textit{\lstinline? run (loop_sched_fact n) (make_state 2 [::n;a] [::] pi_fact) =?}}\\
{\scriptsize \textit{\lstinline? (make_state 14 [::0;(helper_fact n a)] (push (helper_fact n a) [::]) pi_fact)?}}\\
{\scriptsize \textit{\lstinline?Proof.?}}\\
{\scriptsize \textit{\lstinline?move : a; elim : n => [// | n IH a]. ?}}\\
{\scriptsize \textit{\lstinline?by rewrite -IH subn1 -pred_Sn [_ * a]mulnC.?}}\\
{\scriptsize \textit{\lstinline?Qed.?}}\\	
 		\hline
 		\end{tabular}
 	 		}
 	 		
 	\scriptsize{
 		\begin{tabular}{|l|}
 		\hline
 	Exponentiation  \\
 	\hline \hline
 {\scriptsize \lstinline?Lemma program_is_fn_expt n m :  ?}\\
{\scriptsize \lstinline?run (sched_expt n m) (make_state 0 [::n;m] [::] pi_expt) =?}\\
 {\scriptsize \lstinline?  (make_state 14 [::0;fn_expt n m] (push (fn_expt n m) [::]) pi_expt).?}\\
{\scriptsize \lstinline?Proof.?}\\
 {\scriptsize \lstinline?rewrite run_app. by rewrite loop_is_helper_expt.?}\\
{\scriptsize \lstinline?Qed.?}\\
 \\
{\scriptsize \lstinline? Lemma loop_is_helper_expt n m a :?}\\
{\scriptsize \lstinline? run (loop_sched_expt n) (make_state 2 [::n;m;a] [::] pi_expt) =?}\\
{\scriptsize \lstinline? (make_state 14 [::0;(helper_expt n m a)] (push (helper_expt n m a) [::]) pi_expt)?}\\
{\scriptsize \lstinline?Proof.?}\\
{\scriptsize \lstinline?move : n a; elim : m => [// | m IH n a]. ?}\\
{\scriptsize \lstinline?by rewrite -IH subn1 -pred_Sn.?}\\
{\scriptsize \lstinline?Qed.?}\\	
\hline
 		\end{tabular}
 		
 		}
\includegraphics[scale=.2]{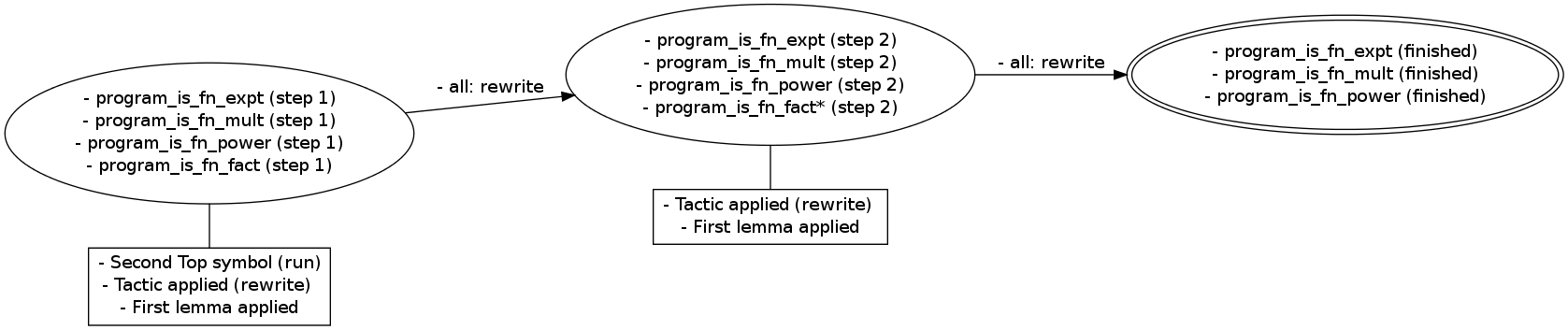}
\caption{{\scriptsize{\textbf{Two top tables. Proofs that the Java bytecode implements the factorial and exponentiation algorithms.}
 	 	\emph{When the user tries to prove \texttt{program\_is\_fn\_fact},
 	he stops after one proof step (top table) and calls ML4PG.  ML4PG suggests a few theorems, like  \texttt{program\_is\_fn\_expt}.
 	It would work for e.g. K-means algorithm and granularity values 
 	from 1 to 4, but using 4 as granularity value the cluster only contains these two lemmas. 
 	In italics, the user reconstructs the proof by analogy with \texttt{program\_is\_fn\_expt} following Proof Strategy~\ref{ps:loop}.} \textbf{Bottom. Automaton-shape representation of the four proofs following Proof Strategy~\ref{ps:loop} and discovered with granularity 3.}
\emph{Note that all the proofs follow the same pattern: the application of the \texttt{rewrite} tactic two consecutive times. Additionally, two of the relevant features that determined the cluster were the lemma applied in the first rewriting tactic (lemma \texttt{run\_app}); and the lemma applied in the second rewriting tactic (a lemma about the loop). Again, notice that ML4PG explicitly suggests which lemmas to analogise!}}}}\label{automaton-jvm2} 
\end{figure}

On the basis of these suggestions, the new team member can try to reconstruct the missing auxiliary lemma and the missing proof steps in the main lemma by analogy. Figure~\ref{automaton-jvm1} shows such analogical reconstruction in italics. This takes him through the first step of the general proof scheme.

%\begin{table}
% 	\centering

% 	\caption{{\scriptsize}}}
% 	\label{tab:jvmproof-stage2}
% \end{table}

%\end{example}

In the second stage, he needs to prove that the Java bytecode implements the factorial algorithm. Again, after a few proof-steps, the user
does not know how to continue the proof, see Figure~\ref{automaton-jvm2}. If ML4PG is invoked at this point, it 
suggests three lemmas (using K-means algorithm and 3 as granularity value) that are used to 
prove that the three Java bytecode programs implement multiplication, exponentiation and power algorithms, respectively. 
All these Java bytecode programs are iterative and involve a loop, and it is easy to notice that the proofs follow the same
proof strategy (see the automaton-shape representation in Figure~\ref{automaton-jvm2}):

\begin{PS}\label{ps:loop}
 
\emph{Prove that the loop implements the helper using an auxiliary lemma. Such a lemma about the loop must consider 
the general case as in the case of Proof Strategy~\ref{ps:spec}. Subsequently, instantiate the result to the concrete case.}
 
\end{PS}

Using this strategy and by analogy with the proofs of the other lemmas of the cluster, the user can finish the proof of lemma 
\lstinline?program_is_fn_fact?; Figure~\ref{automaton-jvm2} shows in italics the reconstruction of that proof. 

\begin{figure}
\centering
  	\scriptsize{
 		\begin{tabular}{|l|}
 		\hline
 Factorial \\
 		\hline
 		\hline
 {\scriptsize \lstinline?Theorem total_correctness_fact n sf :  ?}\\
{\scriptsize \lstinline?sf = run (sched_fact n) (make_state 0 [::n] [::] pi_fact) ->?}\\
 {\scriptsize \lstinline?next_inst sf = (HALT,0\%Z) /\ top (stack sf) = (n`!).?}\\
{\scriptsize \lstinline?Proof.?}\\
 {\scriptsize \lstinline?move => H. split.?}
 {\scriptsize ~~~~~\textit{\lstinline? rewrite H program_is_fn_fact fn_fact_is_theta.?}}\\
{\scriptsize \textit{\lstinline?Qed.?}}\\

 		\hline
 		\end{tabular}
 	 		}
 		
 	\scriptsize{
 		\begin{tabular}{|l|}
 		\hline
 	Exponentiation  \\
 		\hline
 		\hline
{\scriptsize \lstinline?Theorem total_correctness_expt n m sf :  ?}\\
{\scriptsize \lstinline? sf = run (sched_expt m) (make_state 0 [::n;m] [::] pi_expt) ->?} \\
{\scriptsize \lstinline? next_inst sf = (HALT,0%Z) /\ top (stack sf) = (n^m).?} \\
{\scriptsize \lstinline?Proof.?} \\
{\scriptsize ~~~~~~~~~~\lstinline? move => H. split. by rewrite H  program_is_fn_expt fn_expt_is_theta.?}\\
{\scriptsize \lstinline?Qed.?} \\

 		\hline
 		\end{tabular}
 		
 		}
\includegraphics[scale=.22]{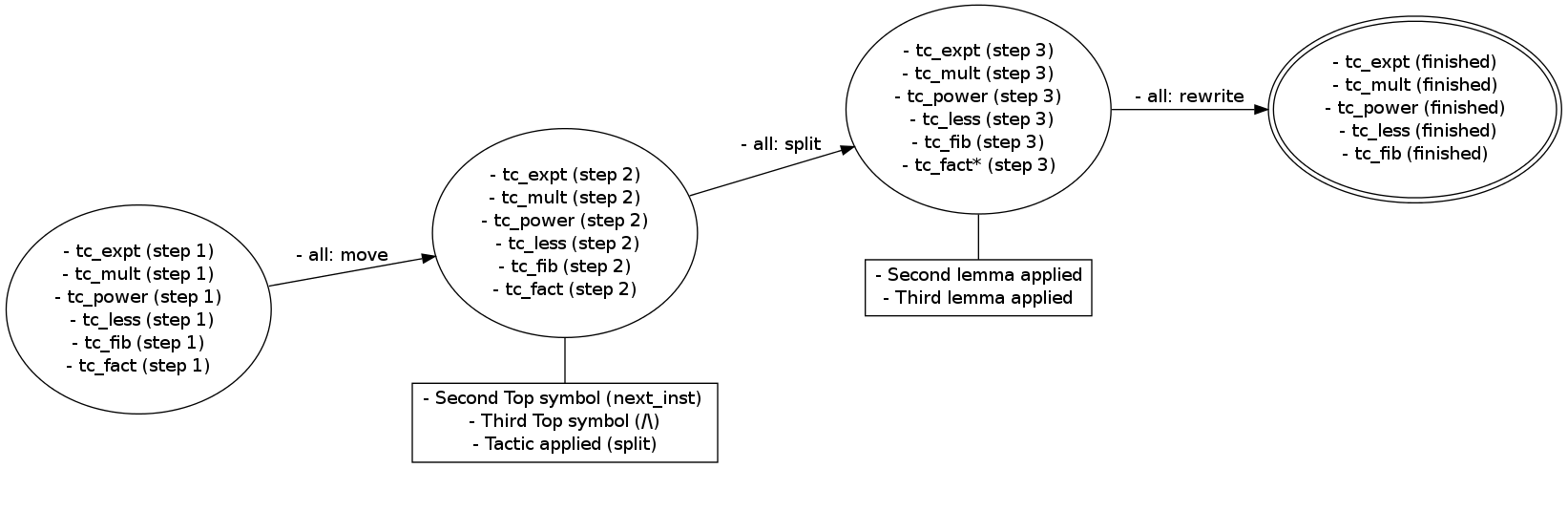}
\caption{{\scriptsize{\textbf{Top. Proofs of total correctness for exponentiation and factorial programs, cf. Theorem~\ref{lem:factorial}.}
 	\emph{The top table shows the initial step to prove Theorem~\ref{lem:factorial} (\texttt{total\_correctness\_fact}).
 	ML4PG suggests a few theorems, in particular,  \texttt{total\_correctness\_expt} is suggested 
	%groups these two lemmas in the same cluster using 
with e.g. K-means algorithm and any granularity values, but using 4 as granularity value gives the cluster containing only these two lemmas. 
 	In italics, the user reconstructs the proof by analogy with the theorem \texttt{total\_correctness\_expt}.}
\textbf{Bottom. Automaton-shape representation of the cluster of six theorems about the total correctness of JVM programs found with  granularity value 3.} \emph{%In the figure, \texttt{tc} stands for \texttt{total\_correctness}. 
Note again that there is always a single transition labelled by the same tactic, showing a strong correlation. In addition, the lemmas applied in the third proof step (lemmas the user would need to analogise) are two of the features that determined the pattern.} }}}\label{automaton-jvm3} 
\end{figure}

Finally, it remains to prove the total correctness of the Java bytecode (Theorem~\ref{lem:factorial}). 
ML4PG finds that the  proofs of the total correctness of $6$ different programs are similar 
and follow the same proof pattern which consists of applying the lemmas obtained from steps $(1)$ and $(2)$, see Figure~\ref{automaton-jvm3}. 
Again, Figure~\ref{automaton-jvm3} illustrates the scenario of calling ML4PG on demand, and using its suggestions to reconstruct the proof by analogy.
Following these guidelines, Theorem~\ref{lem:factorial} can be formalised in Coq/SSReflect by analogy with a similar lemma for e.g. exponentiation, obtaining as a result the proof of the
correctness of the factorial Java bytecode, as shown in Figure~\ref{automaton-jvm3}; see also \cite{HK12} for the full proof.

%\begin{table}
% 	\centering
 	
%  	\caption{{\scriptsize }}}
% 	\label{tab:jvmproofs2}
% \end{table}

The clusters found in the JVM scenario are heterogeneous since they belong to different libraries. 
%However, as in Section~\ref{sec:cas}, we can consider 
%that all the proofs belong to a library about formalisation of JVM code; hence, the clusters are homogeneous. Namely, 
The clusters obtained for the 
different steps %are in the category of clusters that 
consist of lemmas with the same proof structure and use analogous lemmas. This is an interesting kind
of clusters since the analogous lemmas could be automatically generated using techniques presented in~\cite{lpar13}.

\section{ML4PG from the Domain Expert's Perspective}\label{expert}

During the formalisation of a theory, users of interactive theorem provers usually have 
the intuition that there is some kind of similarity or ``\emph{proof-pattern}'' between their current development and other 
theories that they have previously created. The intuition of why two theorems are similar (or 
follow a common proof-pattern) varies from user to user, and is acquired with experience. Therefore, domain expert users with no 
experience using ITPs may have problems to find those similarities. We have seen in the previous scenarios that ML4PG can help to overcome 
this issue and can be used to guide domain expert users in the discovery of patterns in Coq proofs. We consider three kinds of domain expert users of ITPs: novice users of an unexplored domain in ITPs, novice users of a domain previously studied in ITPs, and advanced users; for all these users ML4PG can be helpful.

In the case of domains that have not been previously explored in ITPs, one of the reasons that was in the way for the adoption of ITPs was the lack of libraries with enough background material (for instance, if a domain expert needed a concrete result about sets for his development, he probably had to develop a whole library about set theory from scratch). This problem has been solved as a by-product of projects like the formalisation of the Feit Thompson Theorem~\cite{FTT} or the Flyspeck project~\cite{flyspeck}, and nowadays there are big libraries containing several basic mathematical results. However, these libraries have grown so big that novice users can find difficulties to use them. Imagine that the user wants to explore the SSReflect library of User Scenario 1, inspecting the 1404 theorems one by one is unfeasible; on the contrary, ML4PG provides a snapshot of the library allowing the user to see groups of theorems that are similar. This provides the user a panoramic view of the results 
that are available and what are some of the common techniques used to prove them.

There are some domains where ITP experts have formalised some advanced notions~\cite{formare} and domain experts are willing to extend those libraries with their own results. The advantage for novice users of these domains is that they can use the previous developments as guidance. This situation is similar to User Scenario 3; in this case, the users can use ML4PG not only as a discovery tool of the results that are already formalised, but also to obtain suggestions of similar results that were formalised in the same domain during the construction of a concrete proof.

Finally, domain experts that are advanced users of ITPs can also take advantage of ML4PG features. Even if an advanced user has the intuition that he is proving something similar to one of his previous developments, it can be difficult to remember where he did it. ML4PG can be used to explore the user's libraries and point out where are the proofs that follow a similar pattern to the current proof. This has a positive impact in the formalisation effort since, otherwise, the user should explore his libraries manually (probably re-executing the proofs) to find the theorems that are similar to his current proof-attempt.  

Even if ML4PG can find patterns across different users, it works better finding patterns in the libraries developed by the same user. This is due to the fact that every user has his own style of proving and, therefore, the proof-patterns arise more clearly. This is also the reason why ML4PG works usually better with SSReflect than with plain Coq. SSReflect proofs use only a small number of tactics and this encourages a certain proof style; thanks to this, libraries developed by different people in big developments (e.g. the proof of the Feit Thompson Theorem) should always follow a concrete proof-style, facilitating the discovery of patterns by ML4PG.

% 
% Unlike e.g. related work on using machine-learning for automated premise
% selection~\cite{}
% 
% we do not have a quantitative target when it comes to improving
% interactive proof building experience. No longer speed up in automated proof
% search or the number of automatically proven theorems are the main criteria of
% success. We generally follow the intuition that ML4PG, being an interactive hint
% generator, must provide interesting and non-trivial hints on user’s demand, and
% should be flexible enough to do so at any stage of the proof, and relative to any
% chosen proof library.

\section{Conclusions and Future Work}\label{sec:conclusions}

We have presented three scenarios, of very different nature and from different domains, to test the capabilities of statistical proof-pattern recognition.
We have observed that ML4PG's feature extraction provides sufficiently robust results, tested using  
a few most common clustering algorithms (cf. Table~\ref{tab:jvm}).
%grouping the correct lemmas into clusters, albeit with varied degree of precision. 
Judging by the experiments, the K-means algorithm is the most reliable algorithm,
showing very stable results. The best value for granularity depends on the size of the library, in big libraries (cf. User Scenario  1) granularity values $4$ and
$5$ return the most accurate clusters; however, in small libraries (cf. User Scenarios 2 and 3) the granularity value of $3$ produces better results. 
ML4PG in general requires minimum user effort -- mainly concerning adjustments of the granularity parameter to obtain the result of the required precision. ML4PG is 
very fast and gives instant outputs allowing the user to have quick search/evaluation in an interactive manner.

The most valuable feature of ML4PG is that it works equally well with any library we tried; irrespective of the subject domain or the size of the libraries.
This property can be used to find patterns across subjects, libraries, and users; -- as our case studies illustrate. Moreover, ML4PG discovers two kinds of clusters: homogeneous 
(all the lemmas of the cluster belong to the same library) and heterogeneous (the lemmas of the cluster belong to different libraries). Most of the time, the relation among the
elements of a homogeneous cluster is clear (same proof structure, same lemmas or analogous lemmas). On the contrary, the relation among the elements of a heterogeneous cluster
is more subtle (e.g. a general proof strategy or the use of some kind of auxiliary lemma).  

Work is under way to incorporate the following extensions into ML4PG (see \cite{HK14} for the most recent ML4PG versions):

\begin{itemize}
% 	\item a more sophisticated proof-patch identification for bigger proofs. This paper is based on an ML4PG version that uses only patches of the first few steps in a proof, but see \cite{HK12}
% 	for the experimental version that uses proof-patches to cover entire proofs.
        \item improve the conceptualisation and visualisation of proof-patterns; and,
	\item have a robust data-mining of type declarations and (co-)inductive definitions, alongside the currently used proof-analysis.
\end{itemize}
  
A longer-term project is to generate auxiliary lemmas and definitions automatically, on the basis of statistically discovered patterns. 
We have already done that for ACL2 \cite{lpar13}; however, extrapolation of the techniques of \cite{lpar13} from the first-order untyped 
language of ACL2 to the higher-order dependently-typed language of Coq is a difficult task.

\section*{Acknowledgments}

We would like to thank Marco Gaboardi and Vladimir Komendantsky for proof-reading the paper; their suggestions helped us to improve presentation.
%their comments and suggestions about this paper. 

\bibliographystyle{plain}
\bibliography{biblio}

\begin{thebibliography}{10}

\bibitem{ACDDHPPPT14}
A.~Amorim et~al.
\newblock A verified information-flow architecture.
\newblock In {\em 41st ACM SIGPLAN-SIGACT Symposium on Principles of
  Programming Languages (POPL'14)}, 2014.

\bibitem{AspertiGCTZ04}
A.~Asperti et~al.
\newblock {A Content Based Mathematical Search Engine: Whelp}.
\newblock In {\em Post-Proceedings of the TYPES'04 International Conference},
  volume 3839 of {\em LNCS}, pages 17--32, 2006.

\bibitem{Matita}
A.~Asperti et~al.
\newblock {The Matita interactive Theorem prover}.
\newblock In {\em 23rd International Conference on Automated Deduction
  (CADE'11)}, volume 6803 of {\em LNCS}, pages 64--69, 2011.

\bibitem{ProofGeneral}
D.~Aspinall.
\newblock {Proof General: A Generic Tool for Proof Development}.
\newblock In {\em 6th International Conference on Tools and Algorithms for the
  Construction and Analysis of Systems (TACAS'00)}, volume 1785 of {\em LNCS},
  pages 38--43, 2000.

\bibitem{BB05}
D.~Basin et~al.
\newblock {\em Rippling: Meta-level Guidance for Mathematical Reasoning}.
\newblock Cambridge University Press, 2005.

\bibitem{Ben06}
N.~Benton.
\newblock {M}achine {O}bstructed {P}roof: {H}ow many months can it take to
  verify 30 assembly instructions?, 2006.

\bibitem{BC04}
Y.~Bertot and P.~Cast{\'e}ran.
\newblock {\em Interactive Theorem Proving and Program Development, Coq'Art:
  the Calculus of Constructions}.
\newblock Springer-Verlag, 2004.

\bibitem{Bishop}
C.~Bishop.
\newblock {\em Pattern Recognition and Machine Learning}.
\newblock Springer, 2006.

\bibitem{Blum92}
A.~Blum.
\newblock Learning boolean functions in an infinite attribute space.
\newblock {\em Machine Learning}, 9(4):373--386, 1992.

\bibitem{Agda}
A.~Bove et~al.
\newblock {A Brief Overview of Agda --- A Functional Language with Dependent
  Types}.
\newblock In {\em 22nd International Conference on Theorem Proving in Higher
  Order Logics (TPHOLs'09)}, volume 5674 of {\em LNCS}, pages 73--78, 2009.

\bibitem{BHJM12}
A.~Bundy et~al.
\newblock {AI meets Formal Software Development (Dagstuhl Seminar 12271)}.
\newblock {\em Dagstuhl Reports}, 2(7):1--29, 2012.

\bibitem{Coq}
{\textsc{Coq} development team}.
\newblock {The \textsc{Coq} Proof Assistant Reference Manual, version 8.4}.
\newblock Technical report, 2012.

\bibitem{Duncan02}
H.~Duncan.
\newblock {\em The use of Data-Mining for the Automatic Formation of Tactics}.
\newblock PhD thesis, University of Edinburgh, 2002.

\bibitem{FCT}
G.~Gonthier.
\newblock Formal proof - the four-color theorem.
\newblock {\em Notices of the American Mathematical Society},
  55(11):1382--1393, 2008.

\bibitem{FTT}
G.~Gonthier et~al.
\newblock {A Machine-Checked Proof of the Odd Order Theorem}.
\newblock In {\em 4th Conference on Interactive Theorem Proving (ITP'13)},
  volume 7998 of {\em LNCS}, pages 163--179, 2013.

\bibitem{SSReflect}
G.~Gonthier and A.~Mahboubi.
\newblock {An Introduction to Small Scale Reflection}.
\newblock {\em Journal of Formalized Reasoning}, 3(2):95--152, 2010.

\bibitem{mizar}
A.~Grabowski, A.~Kornilowicz, and A.~Naumowicz.
\newblock {Mizar in a nutshell}.
\newblock {\em Journal of Formalized Reasoning}, 3(2):153--245, 2010.

\bibitem{flyspeck}
T.~Hales.
\newblock {The Flyspeck Project fact sheet}.
\newblock Project description available at
  \url{http://code.google.com/p/flyspeck/}, 2005.

\bibitem{Weka}
M.~Hall et~al.
\newblock {The WEKA Data Mining Software: An Update}.
\newblock {\em SIGKDD Explorations}, 11(1):10--18, 2009.

\bibitem{lpar13}
J.~Heras et~al.
\newblock {Proof-Pattern Recognition and Lemma Discovery in ACL2}.
\newblock In {\em 19th Logic for Programming Artificial Intelligence and
  Reasoning (LPAR-19)}, volume 8312 of {\em LNCS}, pages 389--406, 2013.

\bibitem{HK12}
J.~Heras and E.~Komendantskaya.
\newblock {ML4PG: downloadable programs, manual, examples}, 2012--2014.
\newblock \url{http://staff.computing.dundee.ac.uk/katya/ML4PG/}.

\bibitem{HK14}
J.~Heras and E.~Komendantskaya.
\newblock {Proof Pattern Search in Coq/SSReflect}, 2014.
\newblock \url{http://arxiv.org/abs/1402.0081}.

\bibitem{M03}
{J S. Moore}.
\newblock {\em Models, Algebras and Logic of Engineering Software}, chapter
  {Proving Theorems about Java and the JVM with ACL2}, pages 227--290.
\newblock IOS Press, 2004.

\bibitem{JDB11}
M.~Johansson et~al.
\newblock Conjecture synthesis for inductive theories.
\newblock {\em Journal of Automated Reasoning}, 47(3):251--289, 2011.

\bibitem{KMM00-1}
M.~Kaufmann, P.~Manolios, and {J}~S. Moore, editors.
\newblock {\em Computer-Aided Reasoning: An approach}.
\newblock Kluwer Academic Publishers, 2000.

\bibitem{KHG13}
E.~Komendantskaya et~al.
\newblock {Machine Learning for Proof General: interfacing interfaces}.
\newblock {\em Electronic Proceedings in Theoretical Computer Science},
  118:15--41, 2013.

\bibitem{realsCoq}
R.~Krebbers and B.~Spitters.
\newblock {Type classes for efficient exact real arithmetic in Coq}.
\newblock {\em Logical Methods in Computer Science}, 9(1):1--27, 2013.

\bibitem{ku12}
D.~K{\"u}hlwein et~al.
\newblock Overview and evaluation of premise selection techniques for large
  theory mathematics.
\newblock In {\em 6th International Joint Conference on Automated Reasoning
  (IJCAR'12)}, volume 7364 of {\em LNCS}, pages 378--392, 2012.

\bibitem{formare}
C.~Lange et~al.
\newblock {A Qualitative Comparison of the Suitability of Four Theorem Provers
  for Basic Auction Theory}.
\newblock In {\em Proceedings of Conferences on Intelligent Computer
  Mathematics (CICM'13)}, volume 7961 of {\em LNCS}, pages 200--215, 2012.

\bibitem{JVM}
T.~Lindholm et~al.
\newblock {The Java Virtual Machine Specification: Java SE 7 Edition}, 2012.

\bibitem{Matlab}
MATLAB.
\newblock {\em version 7.14.0 (R2012a)}.
\newblock The MathWorks Inc., Natick, Massachusetts, 2012.

\bibitem{NPW02}
T.~Nipkow et~al.
\newblock {\em Isabelle/HOL - A Proof Assistant for Higher-Order Logic}, volume
  2283 of {\em LNCS}.
\newblock Springer, 2002.

\bibitem{ler07}
S.~Le Roux.
\newblock {Acyclicity and finite linear extendability: a formal and
  constructive equivalence}.
\newblock In {\em 22nd International Conference on Theorem Proving in Higher
  Order Logics (TPHOLs'09)}, Emerging Trends Proceedings, pages 154--169, 2007.

\bibitem{nash}
S.~Le Roux.
\newblock {Acyclic Preferences and Existence of Sequential Nash Equilibria: A
  Formal and Constructive Equivalence}.
\newblock In {\em 20th International Conference on Theorem Proving in Higher
  Order Logics (TPHOLs'07)}, volume 5674 of {\em LNCS}, pages 293--309, 2009.

\bibitem{Ves06}
R.~Vestergaard.
\newblock A constructive approach to sequential nash equilibria.
\newblock {\em Information Processing Letter}, 97:46--51, 2006.

\end{thebibliography}

% \input{clusters}

% ------------------------------------------------------------------------
\end{document}